%% file: PaperForReview.tex
\documentclass[10pt,twocolumn,letterpaper]{article}

\usepackage[pagenumbers]{cvpr} 

\usepackage{graphicx}
\usepackage{lscape,array}
\usepackage{booktabs}
\usepackage{amsmath}
\usepackage{amssymb}
\usepackage{booktabs}
\usepackage{enumerate}
\usepackage{tabularx}
\usepackage{romannum}
\usepackage{booktabs}
\usepackage{caption}
\usepackage{framed}

\usepackage{mdwtab}
\usepackage{multirow}

\usepackage{enumerate}
\usepackage{siunitx} 
\usepackage{babel,blindtext}

\usepackage[pagebackref,breaklinks,colorlinks]{hyperref}

\usepackage[capitalize]{cleveref}
\crefname{section}{Sec.}{Secs.}
\Crefname{section}{Section}{Sections}
\Crefname{table}{Table}{Tables}
\crefname{table}{Tab.}{Tabs.}

\def\cvprPaperID{00054} 
\def\confName{CVPR}
\def\confYear{2024}

\begin{document}


\title{LEAST: ``\underline{L}ocal" t\underline{e}xt-conditioned im\underline{a}ge \underline{s}tyle \underline{t}ransfer}

\author{Silky Singh, Surgan Jandial, Simra Shahid, Abhinav Java\\
Media and Data Science Research (MDSR), Adobe\\
{\tt\small \{silsingh, jandial, sshahid, ajava\}@adobe.com}
}
\maketitle

\input{cvpr2023-author_kit-v1_1-1/latex/00abstract}

\input{cvpr2023-author_kit-v1_1-1/latex/01intro}

\input{cvpr2023-author_kit-v1_1-1/latex/02related_works}

\input{cvpr2023-author_kit-v1_1-1/latex/03method}

\input{cvpr2023-author_kit-v1_1-1/latex/04expts_results}

\input{cvpr2023-author_kit-v1_1-1/latex/05conclusion}


\clearpage
{\small
\bibliographystyle{ieee_fullname}
\bibliography{egbib}
}

\clearpage
\input{cvpr2023-author_kit-v1_1-1/latex/06appendix}

\end{document}

%% file: cvpr2023-author_kit-v1_1-1/latex/00abstract.tex
\begin{abstract}



\noindent Text-conditioned style transfer enables users to communicate their desired artistic styles through text descriptions, offering a new and expressive means of achieving stylization. In this work, we evaluate the text-conditioned image editing and style transfer techniques on their fine-grained understanding of user prompts for precise ``local" style transfer. We find that current methods fail to accomplish localized style transfers effectively, either failing to localize style transfer to certain regions in the image, or distorting the content and structure of the input image. To this end, we develop an end-to-end pipeline for ``local'' style transfer tailored to align with users' intent. Further, we substantiate the effectiveness of our approach through quantitative and qualitative analysis. The project code is available at: \href{https://github.com/silky1708/local-style-transfer}{https://github.com/silky1708/local-style-transfer}

\end{abstract}



%% file: cvpr2023-author_kit-v1_1-1/latex/01intro.tex
\section{Introduction}
\label{sec:intro}

\noindent Text-conditioned style transfer is an exciting area of research, with the potential to provide end users the creative freedom to express abstract, artistic, or texture styles in free-form text. There has been a considerable progress on this front~\cite{clipstyler, styleclip, language_driven, itstyler}, however existing methods tend to impose a uniform style on the entire image based on a target style description. Users might seek distinct styles for various regions within an image, for e.g., by providing a description like ``apply cubism style to the building in the image", the user indicates the preference for `building' in cubism style, while keeping the rest of the image unchanged.

Thus, we introduce a framework that integrates spatial nuances from user-provided style descriptions into the style transfer process. Our proposed end-to-end pipeline comprises two main stages: First, we extract the region in the image that the user wants to stylize (for e.g., ``building"), and the corresponding style (for e.g., ``cubism"). Next, we ground the region in the input image, resulting in a precise segmentation mask. Finally, we adapt CLIP-based loss functions to constrain the style transfer to the identified region in the image, which we call ``local" style transfer. Our method enables fine-grained style transfer by aligning with the user's intent. We corroborate the effectiveness of our method through quantitative and qualitative analysis. Our human preference study indicates that users prefer our method over CLIPstyler~\cite{clipstyler} $\approx97$\% of the time.

%% file: cvpr2023-author_kit-v1_1-1/latex/02related_works.tex
\begin{figure*}[!ht]
\centering
{%
  \includegraphics[clip, width=0.77\textwidth]{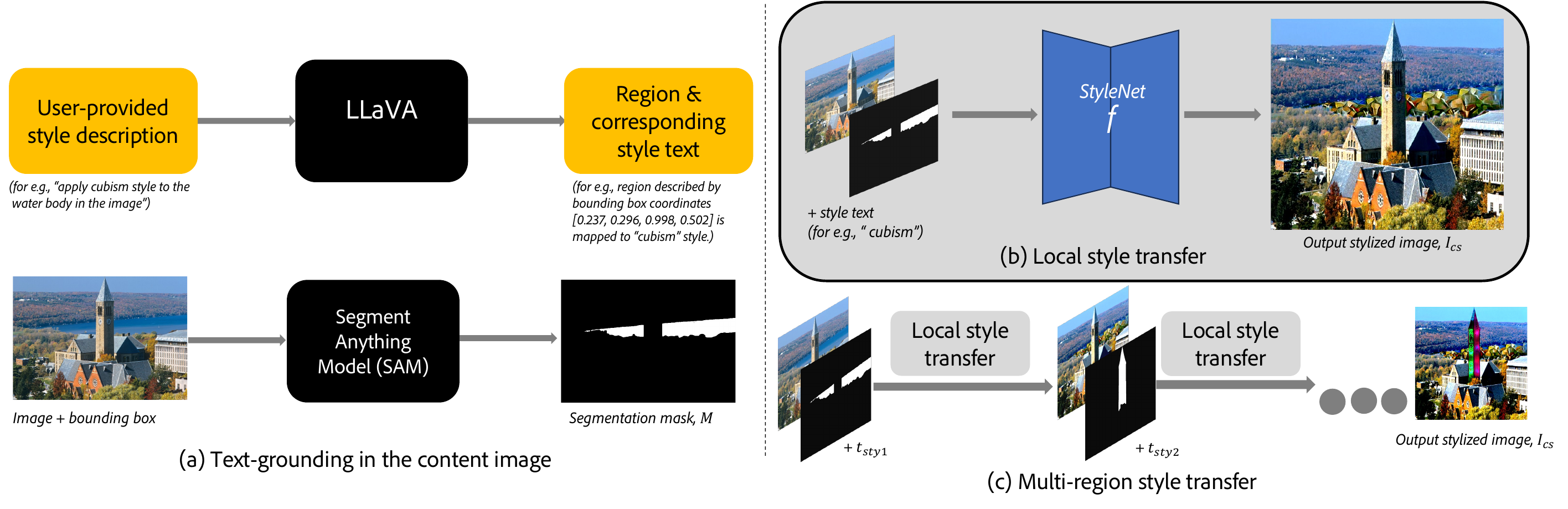}%
}
\caption{Overview of our approach. a) Text-grounding in the content image: We first use LLaVA and SAM to obtain a precise segmentation mask of specified region in the image, and the corresponding desired style. b) Local style transfer: We use the region-style correspondences to constrain style transfer to the specified \textit{local} region. c) This process can be repeated several times to achieve multi-region style transfer.}
\label{fig:architecture}
\end{figure*}

\section{Related Work}
\label{sec:related_works}

\noindent Recent research on \textbf{text-conditioned style transfer}~\cite{language_driven, styleclip, clipstyler} harnesses the multimodal capabilities of VLMs like CLIP~\cite{clip} to stylize an image according to a user-provided description of the desired style. CLVA~\cite{language_driven} leverages a patch-wise style discriminator to jointly embed style images and style instructions. CLIPstyler~\cite{clipstyler} proposes to align the styled image with the style text in CLIP embedding space at both patch and global level. ZeCon~\cite{yang2023zero} is a diffusion-based extension of CLIPstyler. Several modifications to CLIPstyler~\cite{nameyourstyle, kamra2023sem, xu2023stylerdalle} have been proposed to alleviate its limitations, with~\cite{xu2023stylerdalle, nameyourstyle} only focusing on artistic styles, and Sem-CS~\cite{kamra2023sem} solving the over-stylization problem. MOSAIC~\cite{ganugula2023mosaic} is an object-level style transfer method based on CLIPstyler. Additionally, it requires training a BERT-based~\cite{devlin2018bert} text segmentation model. In contrast, ours is an end-to-end inference time optimization with no training required. Following conventional neural style transfer, some methods~\cite{sohn2023styledrop, zhang2023inversion, wang2023stylediffusion} require a style image as an additional input.








Although \textbf{text-conditioned image editing} approaches~\cite{brooks2023instructpix2pix, Geng23instructdiff, bar2022text2live, ge2023expressive, hertz2022prompt, fu2023guiding, parmar2023zero, mokady2023null, kawar2023imagic, goel2023pair} have been quite successful in handling general edit instructions from users, they are not as adept at style transfer. Text2live~\cite{bar2022text2live} relies on explicit region disambiguation by the user to produce alpha masks, which are then used to localize the desired edits. Instruct-pix2pix~\cite{brooks2023instructpix2pix} is a diffusion-based supervised model that follows user's edit instructions. InstructDiffusion~\cite{Geng23instructdiff} further aligns computer vision tasks with human instructions. In practice, these approaches face difficulties in isolating the specified object or region, resulting in unnecessary changes applied to the image.

%% file: cvpr2023-author_kit-v1_1-1/latex/03method.tex
\section{Proposed Methodology}
\label{sec:method}

\noindent For a given input image $I_\text{c}$ (called content image) and a user-provided style description $t_{sty}$ (for e.g., ``apply cubism style to the sky"), we wish to obtain a stylized image $I_\text{cs}$ that respects the semantics of $t_{sty}$. We optimize a style network $f$ that takes $I_\text{c}$ as input, and outputs $I_\text{cs} := f(I_c)$. In the following sections, we elaborate on the building blocks of our method: 1) text grounding in the content image, and 2) local style transfer. Our approach is summarized in Fig. \ref{fig:architecture}.

\begin{figure*}[!th]
    \centering
    \includegraphics[width=0.97\linewidth]{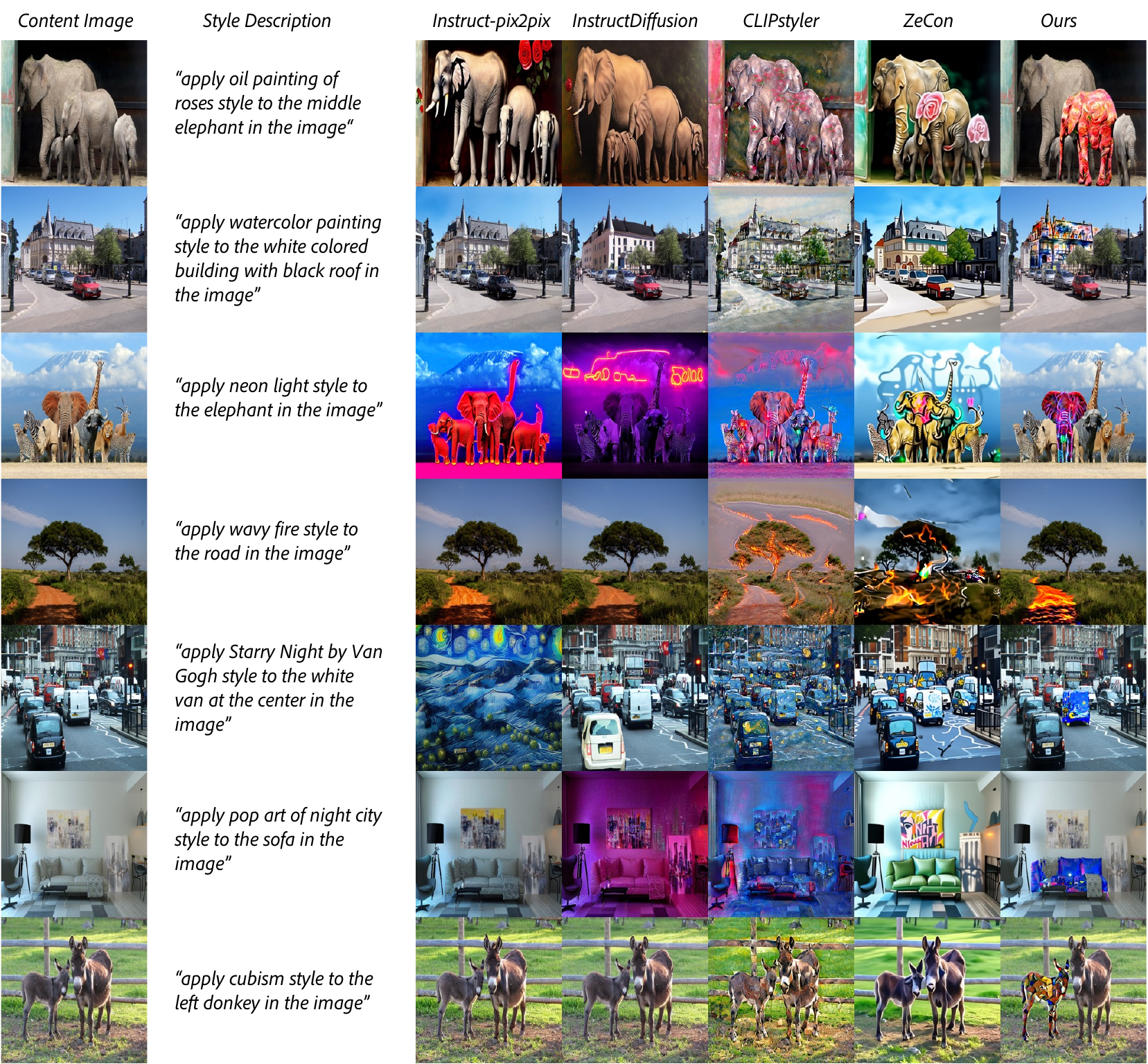}
    \caption{Qualitative comparison of our proposed method with text-conditioned image editing and style transfer approaches. All the baselines fail to localize the desired style transfers, sometimes also failing to preserve the content of the input image (see Instruct-pix2pix $5^{th}$ row). Best viewed in zoom and color. More results in Fig.~\ref{fig:A1_qual1} and \ref{fig:A1_qual2}}
    \label{fig:main_qual_results}
\end{figure*}

\subsection{Text Grounding in the content image}
\label{sec:text_grounding}
\noindent Large vision-language models (VLMs), have been shown to exhibit the remarkable ability to comprehend both image and natural language, and generate contextually relevant responses for a given user query. Building upon its considerable achievements across diverse applications, we propose to query LLaVA~\cite{liu2024llavanext, liu2023improvedllava, liu2023llava}, a large multimodal model, to extract the region and its corresponding style from the user's target style description, $t_{sty}$. In this task, the region, denoted as $R$, refers to a specific area within the content image, such as ``building" or ``sky." The corresponding style, represented as $S$, is the artistic or visual characteristics intended by the user for this region, such as ``Starry night by Van Gogh" or ``cubism." 

In particular, we use LLaVA to extract $R$ (in the form of bounding box coordinates) and $S$, given the content image ($I_c$) and user-provided style description ($t_{sty}$). To achieve this, we query LLaVA using the following prompt: 

\begin{verbatim}
For a given user prompt: '<style desc>',
give the bounding box coordinates of the
object that should be stylized. Also 
return the corresponding style in quotes.
\end{verbatim}

\noindent Here, $<\text{style desc}>$ is used as a placeholder for $t_{sty}$. We parse the output of LLaVA to obtain normalized bounding box coordinates of region $R$ and corresponding style $S$. Subsequently, we utilize these bounding boxes, along with the content image, as inputs to SAM (Segment Anything Model)~\cite{kirillov2023segment} to acquire precise segmentation mask $M$, corresponding to region $R$.

\subsection{Local style transfer}
\noindent In the context of our text grounding approach, we have acquired segmentation mask ($M$) as detailed in \ref{sec:text_grounding}. Our objective is to confine the stylization solely to the identified region in the final output, $I_{cs}$. It is imperative that the regions outside of $R$ remain unaffected and preserve their original appearance. Conventional style transfer methods predominantly concentrate on stylizing the entire image and are unsuitable for \textit{local} style transfer. To alleviate these limitations, we propose to incorporate $M$ in CLIPstyler's~\cite{clipstyler} inference-time optimization:



\begin{enumerate}
    \vspace{-0.15cm}
    
    \item \textbf{Masked CLIP Directional loss ($\overline{L_{dir}}$)}.
    As our focus is to stylize only the region R, we use segmentation mask $M$ to mask both the content and stylized images, and align the direction of CLIP text and image embeddings ($E_T, E_I$ are CLIP text and image encoders respectively). Mathematically, our adapted directional loss can be defined as follows:
    \begin{equation}
    \begin{split}
        \Delta{T} & = E_{T}(S) - E_{T}(t_{src}), ~~t_{src} = \text{``a Photo"} \\
        \overline{\Delta{I}} & = E_{I}(I_{cs} \odot M) - E_{I}(I_{c} \odot M) \\
        \overline{L_{dir}} & = 1 - \frac{\Delta{T}\cdot \overline{\Delta{I}}}{|\Delta{T}||\overline{\Delta{I}}|}
    \end{split}
    \end{equation}
    
    \item \textbf{Masked Patch CLIP loss ($\overline{L_{patch}}$)}. To achieve ``local" stylization within the segmentation mask, we employ a heuristic approach that involves sampling patches from the compact bounding box (say, $B$) around the foreground in $M$. We randomly select $C$ patches of size $p_s \times p_s$ from within this bounding box, for both the content $I_{c,j}$ and the stylized image $I_{cs,j},~j \in \{1,2,3,..., C\}$. The directional CLIP loss ($L_{dir}$) is computed on each patch as follows:
    
    \begin{equation}
        \overline{L_\text{patch}} = \sum_{j=1}^{C} {L_{dir,j}}
    \end{equation}

    \item \textbf{Masked Content Loss ($\overline{L_\text{content}}$)}. To preserve the contents of the image specifically in the region $R$, we crop and resize both the content and stylized images  using the bounding box $B$ to obtain ``local" crops, say $\hat{I_c}$ and $\hat{I_{cs}}$. Our modified content loss is the mean squared error between VGG~\cite{simonyan2014very} features of these crops:
    \begin{equation}
        \overline{L_\text{content}} = || \text{VGG}(\hat{I_c}) - \text{VGG}(\hat{I_{cs}}) ||_2
    \end{equation}

    \item \textbf{Masked Total Variation Loss ($\overline{L_\text{tv}}$)}. Total variation loss is used to promote smoothness in the output image by penalizing abrupt changes or noise. We compute a Hadamard product of the styled image $I_{cs}$ with the segmentation mask $M$. This effectively isolates the stylized region $R$, thus enabling local regularization in the stylized output.
    \begin{equation}
        \overline{L_{tv}} = L_{tv}(I_{cs} \odot M)
    \end{equation}

\end{enumerate}

\noindent We optimize for the following overall objective function to obtain $I_{cs}$ corresponding to style $S$ and region $R$:
\begin{equation}
    \overline{L_\text{total}} = \overline{\lambda_d} \overline{L_\text{dir}} + \overline{\lambda_p} \overline{L_\text{patch}} + \overline{\lambda_c} \overline{L_\text{content}} + \overline{\lambda_{tv} L_{tv}}
\end{equation}

\noindent After the local optimization, we obtain the final output by copying the background region in $M$ from the original content image, $I_c$. Note that we use the same notation $I_{cs}$ to denote the outputs before and after this operation. \begin{equation}
    I_{cs} := I_{cs} \odot M + I_c \odot (1 - M)
\end{equation}
    

\noindent \textbf{Multi-region style transfer}. Our end-to-end local style transfer process can be systematically iterated for a sequence of regions $R_1, R_2, ..., R_N$. The final stylized output $I_{cs}$ is derived from the last local style transfer operation corresponding to region $R_N$. This iterative approach ensures that each region is stylized individually, and the resulting image reflects the cumulative effect of all the local style transfers. Refer Fig.~\ref{fig:multi_region_results}

\begin{equation}
    I_{cs} = f(...f(f(I_c, M_1), M_2), ..., M_N)
\end{equation}

%% file: cvpr2023-author_kit-v1_1-1/latex/04expts_results.tex
\section{Experiments and Results}
\label{sec:experiments}

\noindent \textbf{Dataset}. We curate a set of 25 natural images, spanning broad categories including interior, outdoors/street, natural scenery, animals, inanimate objects. For each image in our dataset, we manually design 10 prompts with a format of ``apply $<\text{style}>$ style to $<\text{region}>$ in the image," where $<\text{style}>$ is a placeholder for artistic or texture styles, for e.g., ``oil painting of roses", ``white wool", ``watercolor painting", ``Starry Night by Van Gogh" etc. and $<\text{region}>$ is a placeholder for an object/region in the image. In case of multiple objects of same category, we use prepositions like ``the leftmost", ``at the center" etc. to resolve ambiguities. Note that there is no standard dataset for our task, and since our approach is an end-to-end inference-time optimization, we only use this dataset for evaluation purposes.

\vspace{0.2cm}
\noindent \textbf{Model architecture \& hyperparameters}. The neural network $f$ is a U-Net~\cite{unet} architecture with 3 downsample and 3 upsample layers, with channel dimensions of 16, 32, 64 respectively. Following~\cite{gatys}, we use the VGG~\cite{simonyan2014very} features from layers ``conv4\_2" and ``conv5\_2" to compute the content loss. The number of crops for Patch CLIP loss, $C$ is set to 64 and patch size $p_s = 100$. We interpolate the content images to a resolution of $512 \times 512$ for stylization, except ZeCon where the resolution is $256\times256$. The loss coefficients are set as follows: $\overline{\lambda_d} = 500,~\overline{\lambda_p} = 10^{3},~\overline{\lambda_c} = 150, ~\overline{\lambda_{tv}} = 2\times10^{-3}$. We use an Adam optimizer with a learning rate of $5\times 10^{-4}$. Empirically, the network $f$ converges in $\approx$200 iterations.

\vspace{0.2cm}
\noindent \textbf{Baselines}. We compare our approach against two broad sets of baselines: text-conditioned image editing (Instruct-pix2pix\cite{brooks2023instructpix2pix}, InstructDiffusion\cite{Geng23instructdiff}), and text-conditioned image style transfer (CLIPstyler\cite{clipstyler}, ZeCon\cite{yang2023zero}) techniques.



\vspace{0.2cm}
\noindent \textbf{Evaluation}. CLIP score~\cite{clip} is a common evaluation metric to quantify the alignment of an image with a given text. To effectively compute CLIP scores for local style transfer task, we first mask the local region $R$ using $M$ as follows: $I_{cs} \odot M$, and then crop using the compact bounding box $B$: crop$_B$($I_{cs} \odot M$). We report the CLIP score between this crop and style $S$ in Tab~\ref{tab:main_quant_res}. In practice, CLIP score does not correlate well with human preferences~\cite{wu2023human}. We conduct a user study to circumvent this issue. We ask $\approx$20 users to choose between CLIPstyler and our outputs for given pairs of content images ($I_c$) and style descriptions ($t_{sty}$). The results are averaged across users and reported in Tab~\ref{tab:user_study}.


\vspace{0.2cm}
\noindent \textbf{Results}. While Table~\ref{tab:main_quant_res} shows that our method is comparable to baselines in CLIP score, it outperforms the highest scoring CLIPstyler in the user study (Table~\ref{tab:user_study}) by a large margin. CLIPstyler and other style transfer techniques tend to stylize the entire image with no localization (refer Fig.~\ref{fig:main_qual_results}). On the other hand, image editing methods tend to distort the contents of the input image, with little to no style transfer. Qualitative results are shown in Fig.~\ref{fig:main_qual_results}.

\begin{table}
\centering
\footnotesize
\begin{tabular}{lc}
\toprule
Methods       & CLIP score ($\uparrow$) \\
\midrule
Instruct-pix2pix~\cite{brooks2023instructpix2pix} & 21.58 \\
InstructDiffusion~\cite{Geng23instructdiff} & 21.58 \\
CLIPstyler~\cite{clipstyler} & \textbf{21.64} \\
ZeCon~\cite{yang2023zero} & 21.54 \\
Ours & 21.47 \\
\bottomrule
\end{tabular}
\caption{Quantitative comparison of our method against baselines, measured by CLIP score.}
\label{tab:main_quant_res}
\end{table}

\begin{table}
\centering
\footnotesize
\begin{tabular}{lcc}
\toprule
     & CLIPstyler~\cite{clipstyler} & Ours \\
\midrule
Avg. preference ($\uparrow$) & 2.11\% & \textbf{97.89\%} \\
\bottomrule
\end{tabular}
\caption{Human preference study with CLIPstyler and our method.}
\label{tab:user_study}
\end{table}

%% file: cvpr2023-author_kit-v1_1-1/latex/05conclusion.tex
\section{Conclusion}
\label{sec:conclusion}

\noindent In this paper, we identify the limitations of existing text-conditioned style transfer and image editing approaches in failing to perform local stylization. To this end, we propose an effective pipeline that involves text grounding to achieve fine-grained local style transfer. We hope this work motivates further research into style transfer applications.



%% file: cvpr2023-author_kit-v1_1-1/latex/06appendix.tex



\begin{figure*}[!hbt]
    \includegraphics[width=1.0\linewidth]{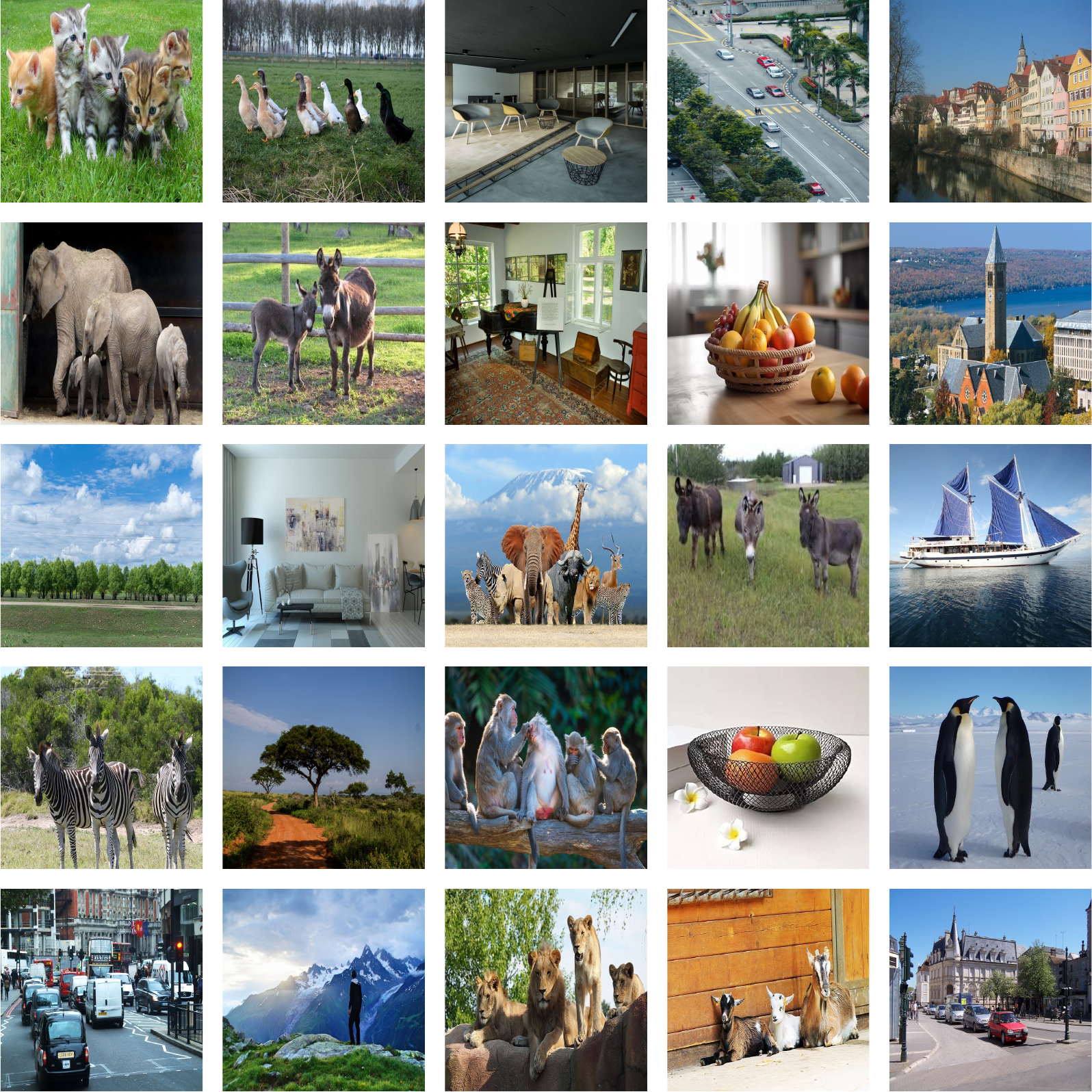}
    \caption{Snapshot of our dataset. We collect a set of 25 natural images to evaluate the efficacy of various approaches on the task of local style transfer. The copyrights exist with respective owners of these images.}
    \label{fig:dataset_snapshot}
\end{figure*}

\clearpage
\begin{figure*}[]
    \centering
    \includegraphics[width=0.9\linewidth]{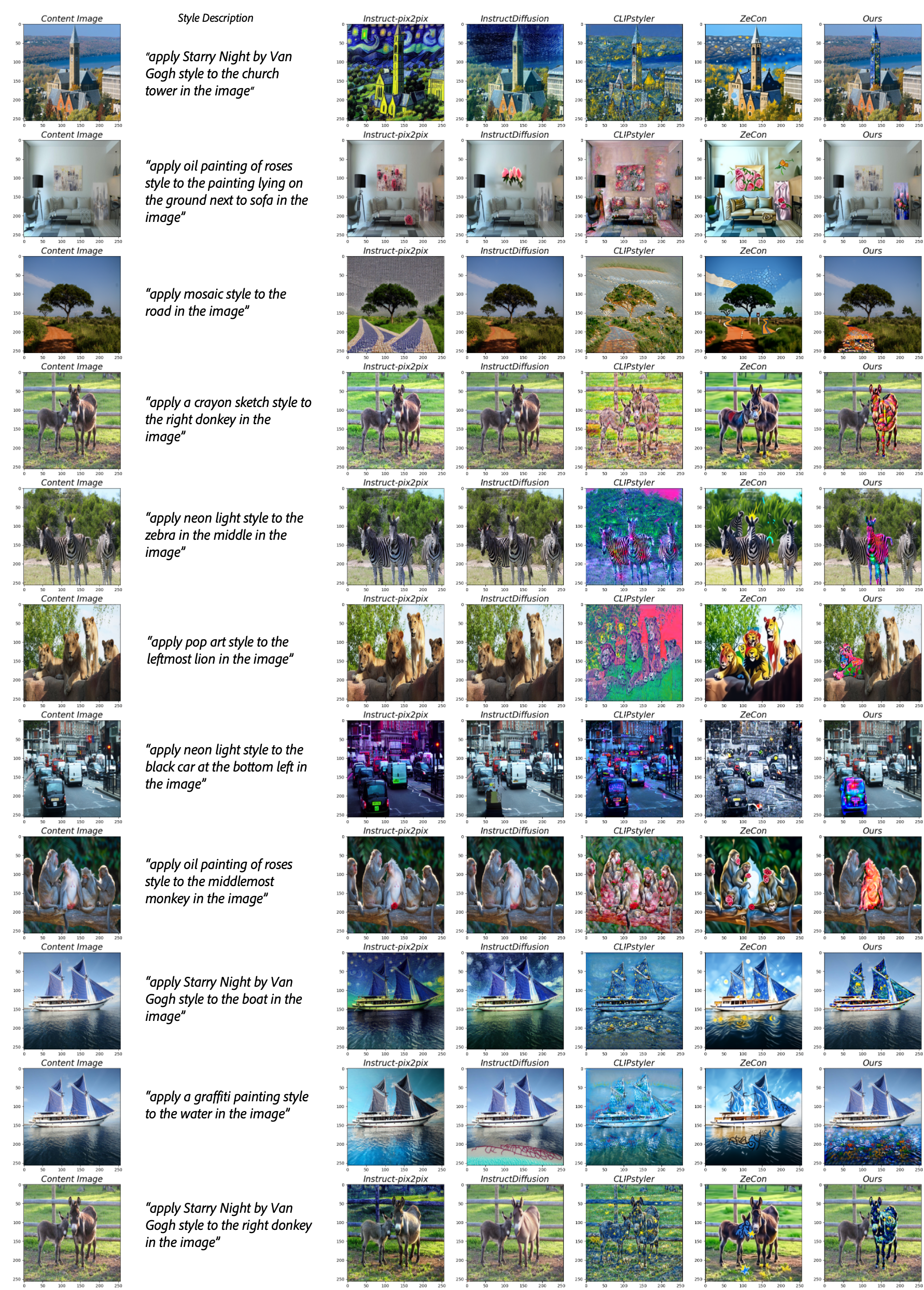}
    \caption{Qualitative comparison of our proposed method with text-conditioned image editing and style transfer approaches.}
    \label{fig:A1_qual1}
\end{figure*}

\clearpage
\begin{figure*}[]
    \centering
    \includegraphics[width=0.9\linewidth]{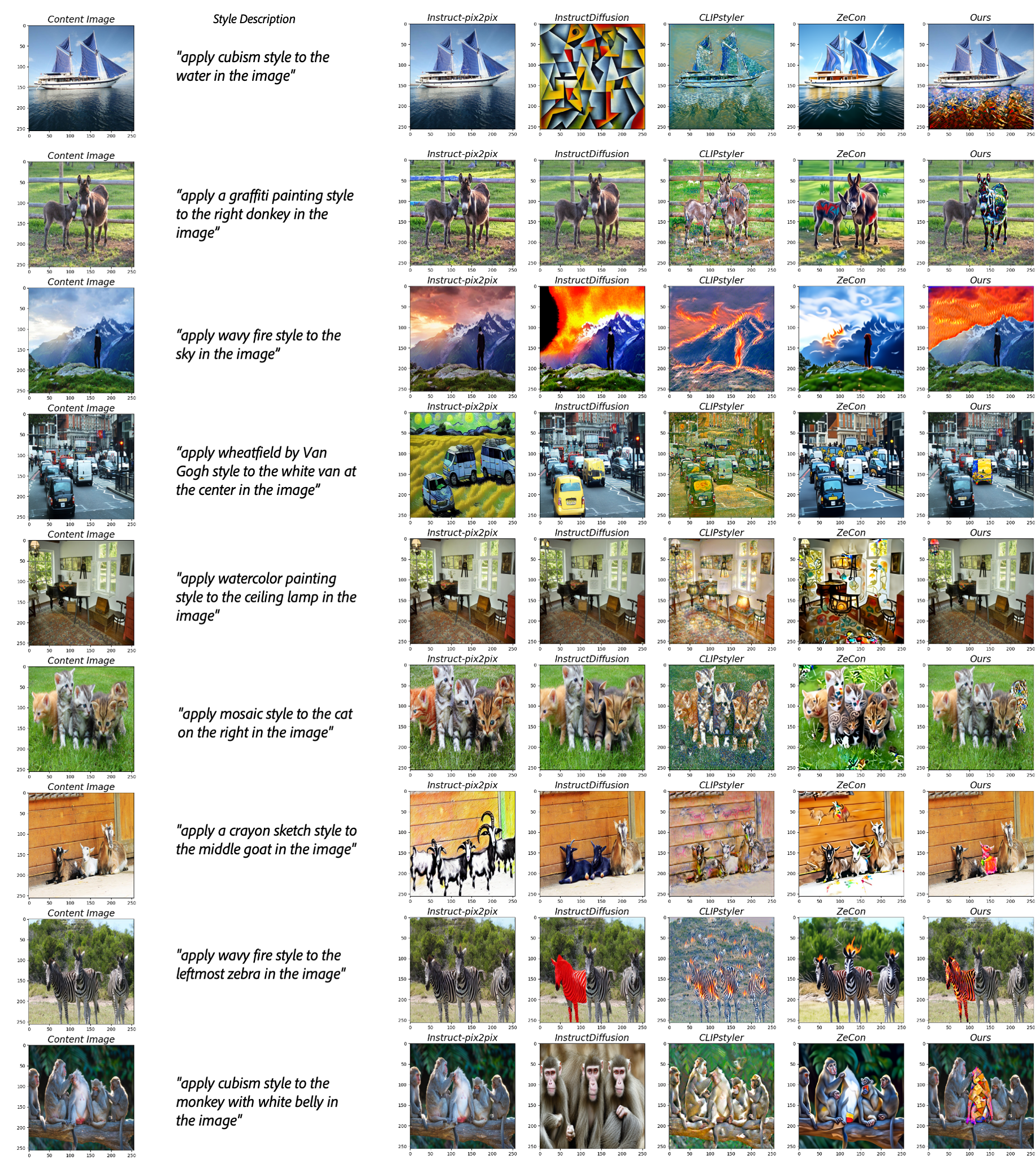}
    \caption{Qualitative comparison of our proposed method with text-conditioned image editing and style transfer approaches.}
    \label{fig:A1_qual2}
\end{figure*}

\clearpage
\begin{figure*}[]
    \centering
    \includegraphics[width=1.0\linewidth]{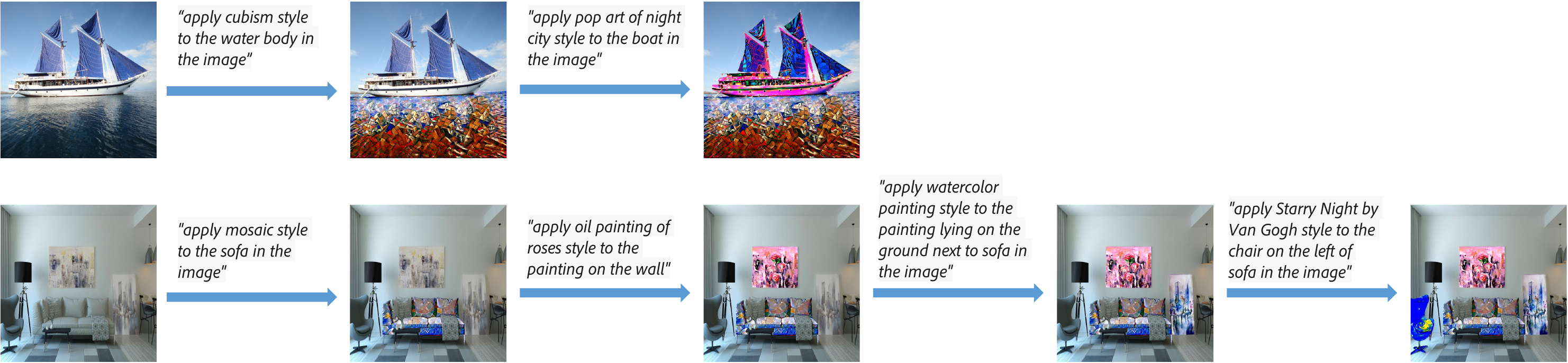}
    \caption{Results for multi-region style transfer using our method. Our \textit{local} style transfer approach can be applied sequentially to any number of regions with different style descriptions to obtain the final stylized image.}
    \label{fig:multi_region_results}
\end{figure*}

\begin{figure*}[]
    \centering
    \includegraphics[width=1.0\linewidth]{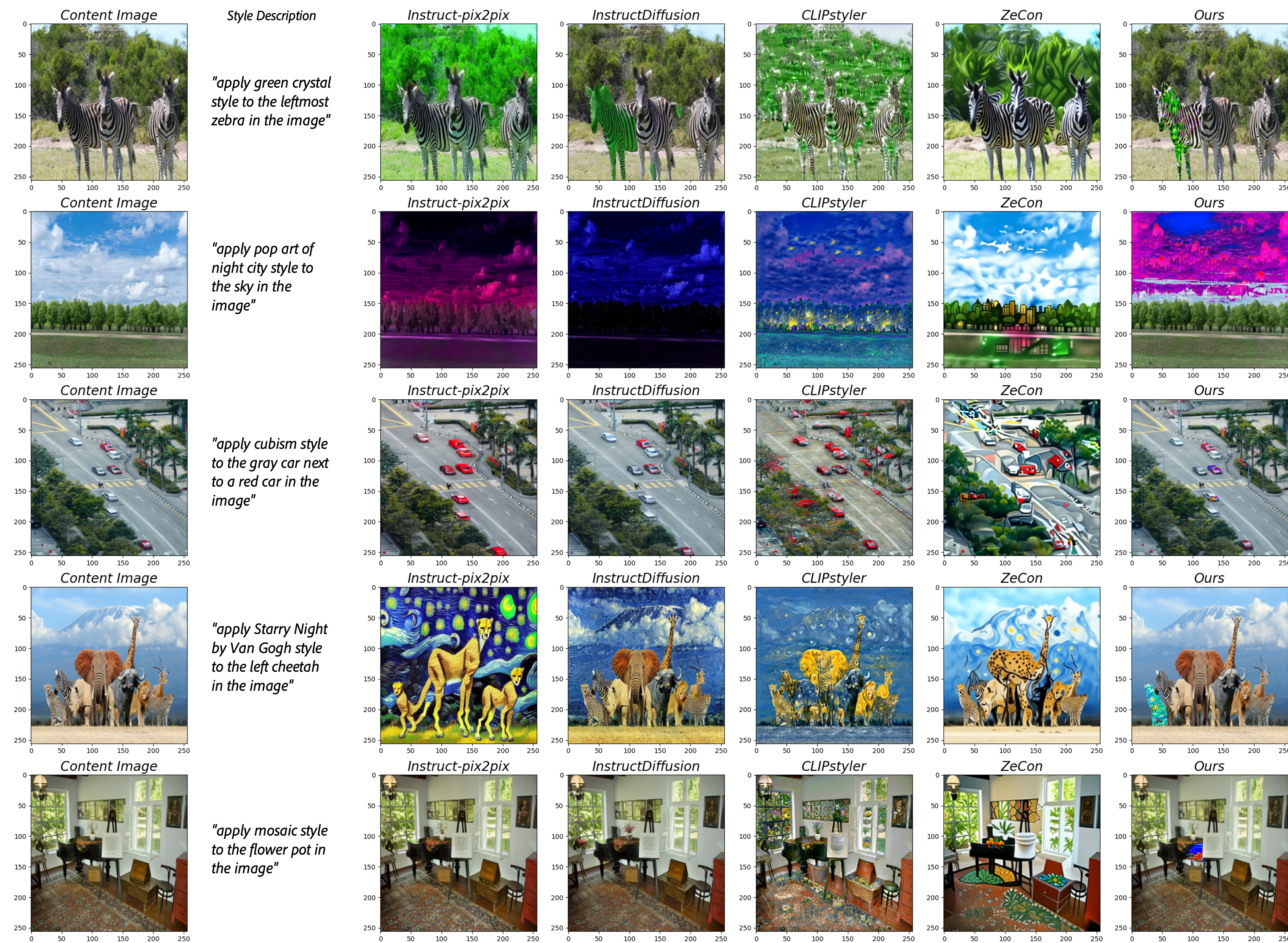}
    \caption{Some failure cases of our approach. Our proposed method struggles when the target object is too small, or the segmentation of the target region in the image (Sec.~\ref{sec:text_grounding}) is not precise.}
    \label{fig:failure_cases}
\end{figure*}